\documentclass{article}
\usepackage[numbers]{natbib}

\usepackage[final]{neurips_2024}

\usepackage[utf8]{inputenc} 
\usepackage[T1]{fontenc}    
\usepackage{hyperref}       
\usepackage{url}            
\usepackage{booktabs}       
\usepackage{amsmath, amsfonts}       
\usepackage{nicefrac}       
\usepackage{microtype}      
\usepackage{xcolor}         
\usepackage{graphicx}       
\usepackage[capitalize,noabbrev]{cleveref}
\usepackage{siunitx}
\crefname{figure}{Fig.}{Figs.}
\definecolor{kleinblue}{RGB}{0, 47, 167} 
\definecolor{kleinblue2}{RGB}{20, 20, 125} 
\definecolor{kleinred}{HTML}{bc1919}

\title{Pretraining Frequency Predicts Compositional Generalization of CLIP on Real-World Tasks}

\author{
    Thaddäus Wiedemer$^{\,1,2,3}$\footnotemark[1] \quad Yash Sharma$^{\,1,2,3}$\footnotemark[1] \quad Ameya Prabhu$^{\,2,3}$ \\
    \\
    \textbf{Matthias Bethge}$^{\,2,3,4}$ \quad \textbf{Wieland Brendel}$^{\,1,2,4}$\\
    \\
    $^{1\,}$Max-Planck-Institute for Intelligent Systems \quad $^{2\,}$Tübingen AI Center\\
    $^{3\,}$University of Tübingen \quad $^{4\,}$ELLIS Institute Tübingen\\
    \\
    {\tt\small thaddaeus.wiedemer@gmail.com, ysharma1126@gmail.com}
}

\begin{document}
\renewcommand*{\thefootnote}{\fnsymbol{footnote}}
\footnotetext[1]{Equal contribution.}

\maketitle

\begin{abstract}
We investigate the success conditions for compositional generalization of CLIP models on real-world data through performance prediction.
Prior work shows that CLIP requires exponentially more pretraining data for linear performance gains on individual concepts.
This sample-inefficient scaling could be mitigated if CLIP systematically understood new inputs as compositions of learned components, allowing rare observation to be mapped to common concepts.
To explore CLIP's compositional generalization ability, we filter retrieval corpora for samples with object combinations not present in the pretraining corpus.
We show that CLIP's performance on these samples can be accurately predicted from the pretraining frequencies of individual objects.
Our findings demonstrate that CLIP learns to disentangle objects observed in its pretraining data and can recompose them straightforwardly.
Additionally, we are the first to show how this ability scales with pretraining data.
For data curation in practice, our results suggest that balancing object occurrences improves generalization, which should benefit CLIP's efficiency and accuracy without scaling data volume.
\end{abstract}

\renewcommand*{\thefootnote}{\arabic{footnote}}
\setcounter{footnote}{0}

\section{Introduction}
Vision Language Models (VLMs) like CLIP~\citep{radford2021learning} have seen widespread adoption for downstream tasks like classification, image retrieval, and image generation due to their transferability and impressive zero-shot performance.
However, CLIP models are data-hungry, requiring exponentially more pretraining data for linear performance gains on downstream samples~\citep{udandarao2024zeroshotexponentialdatapretraining}.
Similar sample-inefficient scaling has been reported for Large Language Models (LLMs)~\citep{kandpal2023large, antoniades2024generalization}, raising doubts about the feasibility of improving zero-shot performance of foundation models by scale alone.

A systematic way to overcome inefficient scaling is thought to be compositional generalization---the ability to understand and form novel combinations of learned concepts~\citep{fodor1988connectionism,hupkes2020compositionality,wiedemer2023compositional}.
A model that can generalize in this way should more effectively combine learned concepts to understand new inputs, ultimately leading to increased zero-shot performance.
Yet, the compositional abilities of large VLMs, such as CLIP, remain poorly understood:
Existing studies in the visual domain are either theoretical, operate on synthetic data, or fail to verify whether compositions used for evaluation are truly novel given the pretraining data (see Sec.~\ref{sec:related}).
Moreover, \emph{the relationship between a VLM's compositional abilities and its pretraining data is entirely uncharacterized}.

To address this gap, we aim to investigate CLIP's compositional generalization on real-world data as a function of its pretraining corpus.
Specifically, we make the following contributions:
\begin{itemize}
    \item We leverage the scalable concept-extraction pipeline proposed by \citet{udandarao2024zeroshotexponentialdatapretraining} to curate text-to-image (T2I) and image-to-text (I2T) retrieval test sets for compositional generalization (Sec.~\ref{sec:compgen_filter}). Each test set is created for the pretraining corpus of the tested model such that it contains only samples with novel combinations of known object classes.
    \item We show that CLIP models perform consistently well on our curated test sets, regardless of the architecture or scale of the backbone.
    \item We demonstrate that across architectures, parameter counts, and pretraining data scales, CLIP's ability to compose objects can be accurately predicted from the independent pretraining frequencies of each object in the composition (Sec.~\ref{sec:results}).
\end{itemize}

Our results are the first to establish a firm connection between the compositional generalization of a VLM and its pretraining data. The nature of this connection shows that CLIP obtains an independent understanding of object classes from web-scale data.

\section{Related Work}\label{sec:related}

\paragraph{Theoretical Works \& Synthetic Data}
A growing body of works~\citep{montero2021role,monteroLostLatentSpace2022,schott2022visual,lewis2022does,wiedemer2023compositional,wiedemer2024provable,okawaCompositionalAbilitiesEmerge2023, jung2024learning} provides significant theoretical understanding of compositional generalization results in the vision domain.
Similar works exist for compositionally in language~\citep{fodor1988connectionism,hupkes2020compositionality,berlot2023attribute}, often under the more specific term \emph{systematicity}~\citep{fodor1988connectionism,hupkes2020compositionality,berlot2023attribute}.
In the language domain, promising progress has been made~\citep{lake2023human}, but results in both domains nonetheless remain confined to synthetic datasets~\citep{lake2018generalization,kim2020cogs}l; \citet{sun-etal-2023-validity} actively questions the transferability of insights to real-world data. In contrast, our work analyzes compositional generalization using real-world retrieval datasets.

\paragraph{VLM Benchmarks \& Contamination}
Several compositionality benchmarks have been proposed for VLMs~\citep{thrush2022winoground, lewis2022does, zhao2022vl, yuksekgonul2023when, ma2023crepe, hsieh2023sugarcrepe, ray2024cola, wangEnhancingCompositionalGeneralization2024, abbasi2024deciphering}. However, these studies do not consider the overlap of concept combinations with web-scale pretraining data.
Data contamination of this kind has been shown to significantly impact CLIP's zero-shot performance~\cite{mayilvahanan2024search}, making it difficult to distinguish between genuine generalization and mere memorization.
A notable exception is \citet{abbasi2024deciphering}, who generate test images of novel attribute-object pairs, but as a result, their benchmark resorts to synthetic data. Our work controls for data contamination by only considering combinations that do not occur in the pretraining data but do occur in real-world benchmarks.

\section{Predicting Compositional Generalization from Pretraining Frequency}\label{sec:experiments}
We adapt the pipeline from \citet{udandarao2024zeroshotexponentialdatapretraining} in two steps to study the success conditions for compositional generalization.
First, we use it to curate retrieval test sets that contain novel combinations of objects with respect to a pretraining set (Sec.~\ref{sec:compgen_filter}). Second, we propose a simple modification to predict downstream performance in terms of samples rather than concepts (Sec.~\ref{sec:f_sample}). Finally, we evaluate CLIP models with varying architectures, parameter counts, and pretraining data scales and show consistent scaling behavior (Sec.~\ref{sec:results}).

\subsection{Curating Compositional Generalization Test Sets}\label{sec:compgen_filter}
We consider two standard retrieval datasets: Flickr-1K~\citep{young2014image} and COCO-5K~\citep{lin2014microsoft}. Both can be used for benchmarking text-to-image (T2I) or image-to-text (I2T) retrieval.

To measure compositional generalization, we follow \citet{hupkes2020compositionality} and retain only test samples containing multiple concepts $o_1, ..., o_n$, where
\renewcommand{\labelenumi}{(\roman{enumi})}
\begin{enumerate}
    \item the model has been familiarized with each concept $o_i$ only in the absence of $o_{j \neq i}$,
    \item the combination $o_1, ..., o_n$ is plausible.
\end{enumerate}

\citet{udandarao2024zeroshotexponentialdatapretraining} compile a list of 945 nouns in the text captions for these 2 retrieval datasets as possible concepts.
The presence of a concept in a pretraining sample is established if it is part of the caption (after lemmatization) \emph{and} it is found in the image using RAM++~\citep{huang2023open}.
Consequently, our analysis of concepts is limited to tangible objects; more abstract concepts like actions or stylistic information are harder to annotate in the visual domain and can, therefore, not be reliably quantified using this pipeline.
With this setup, the frequency $f_{\mathcal D}(o)$ of object class $o$ in a pretraining corpus $\mathcal D$ simply counts the number of samples it occurs in.\footnote{We use the term frequency instead of count since we only compare quantities for a given, fixed-size pretraining set, in which case normalization can be omitted.}

To address (i), we first consider how often objects $o_1, ..., o_n$ in a test sample $x$ co-occur in the pretraining dataset. We call this quantity the co-occurrence frequency, formally given by
\begin{equation}\label{eq:f_cap}
    f_{\cap, \mathcal D}(x) = \left\| \left\{d \in \mathcal D \;|\; \text{for all } o \in x: o \in d \right\} \right\|.
\end{equation}
Condition (i) above is then satisfied by test samples $x$ which contain a novel combination of objects, i.e., $f_{\cap, \mathcal D}(x) = 0$, but each object has been observed at least once, i.e., $f_{\mathcal D} > 0$ for all $o \in x$.

Condition (ii) is hard to verify in general but is trivially met here since we filter real-world data.

Since all frequencies are dependent on the pretraining corpus, the size of our compositional generalization test sets differ. The number of samples in each test set (total and percentage) is shown in Figs.~\ref{fig:lines_t2i_k10}~and~\ref{fig:lines_i2t_k10}.

\subsection{Per-Sample Prediction}\label{sec:f_sample}
\paragraph{Metrics}
We measure performance on each sample using Recall@$k$ for $k \in \{1, 5, 10\}$ following \citet{radford2021learning}. Figs.~\ref{fig:lines_t2i_k10}~and~\ref{fig:lines_i2t_k10} show results for Recall@10, other results are listed in App.~\ref{app:extra_plots}.

\paragraph{Sample Frequency}
We compute the average pretraining frequency $f_\text{avg}$ of each test sample $x$ as the geometric mean of the frequencies of the objects $o_1, ..., o_n$ in the sample $x$, i.e.,
\begin{equation}
    f_\text{avg}(x) = \left(\prod_{o \in x}^n f(o)\right)^\frac{1}{n}.
\end{equation}
The choice of geometric mean is motivated by the assumption that the model's performance on a combination of objects depends on the quality of the model's independent understanding of each object in the combination. For example, a simple retrieval engine might find samples containing two objects $o_1, o_2$ without considering their interaction by first finding samples containing $o_1$ and then filtering the results for samples containing $o_2$. In this case, the probability of retrieving a correct sample based on the prompt $\mathcal P$ can be written as
\begin{equation}
    P(y=1 | o_1, o_2 \in \mathcal P) = P(y=1 | o_1 \in \mathcal P) P(y=1 | o_2 \in \mathcal P).
\end{equation}
The geometric mean reflects the multiplicative impact of each object on the whole~\citep{okawaCompositionalAbilitiesEmerge2023}.

\paragraph{Fitting a Predictor}
Our evaluation yields $(y, f_\text{avg})$ for each test sample, where $y=1$ (0) indicates correct (wrong) retrieval. For each test set, we drop noisy outliers via IQR-removal on $f_\text{avg}$, following \citet{kandpal2023large, udandarao2024zeroshotexponentialdatapretraining}. We fit a logistic regression model with bootstrapping to predict $P(y = 1)$ given $f_\text{avg}$.

\subsection{Results}\label{sec:results}
\begin{figure}
    \centering
    \includegraphics[width=\linewidth]{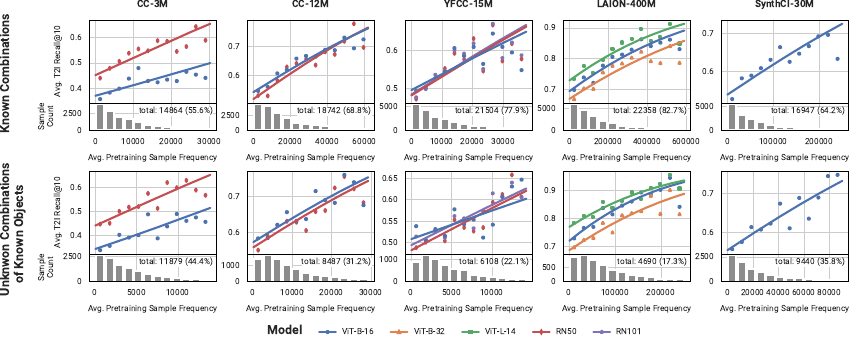}
    \caption{\textbf{T2I Recall@10}. CLIP's performance on unknown combinations (bottom) matches that on known combinations (top) and can be consistently predicted as a linear function of the average pretraining frequency of the constituent objects. All regression fits are significant at $p<0.01$.}
    \label{fig:lines_t2i_k10}
\end{figure}

\begin{figure}
    \centering
    \includegraphics[width=\linewidth]{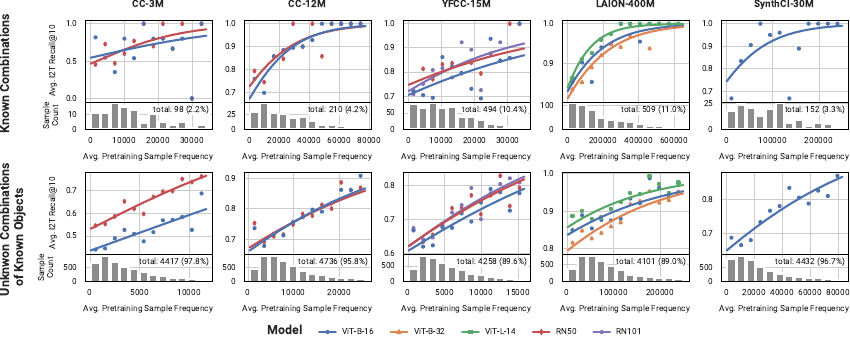}
    \caption{\textbf{I2T Recall@10}. CLIP's performance on unknown combinations (bottom) almost matches that on known combinations (top) and can be consistently predicted as a linear function of the average pretraining frequency of the constituent objects. All regression fits are significant at $p<0.01$.}
    \label{fig:lines_i2t_k10}
\end{figure}

Fig.~\ref{fig:lines_t2i_k10} collects results for the T2I task, Fig.~\ref{fig:lines_i2t_k10} for I2T. More results can be found in App.~\ref{app:extra_plots}.

\paragraph{Models}
We test CLIP~models~\citep{radford2021learning} with both ResNet~\citep{he2016deep} and Vision~Transformer~\citep{dosovitskiy2020image} architectures. Specifically, we evaluate ViT-B-16~\citep{mu2022slip} and RN50~\citep{goel2022cyclip,nguyen2022quality} trained on CC-3M~\citep{sharma2018conceptual} and CC-12M~\citep{changpinyo2021conceptual}; ViT-B-16, RN50, and RN101~\citep{ilharco2021openclip} trained on YFCC-15M~\citep{thomee2016yfcc100m}; ViT-B-16, ViT-B-32, and ViT-L-14 trained on LAION400M~\citep{schuhmann2021laion}; and ViT-B-16 trained on SynthCI-30M~\citep{hammoud2024synthclip}. We follow \texttt{open\_clip}~\citep{ilharco2021openclip}, \texttt{slip}~\citep{mu2022slip} and \texttt{cyclip}~\citep{goel2022cyclip} for implementation details.

\paragraph{Fraction of Compositional Generalization Samples}
Looking at the histogram percentages, \SI{17}-\SI{44}{\%} of samples are unknown compositions of known concepts for the T2I task. For I2T, the fraction is much higher with \SI{89}-\SI{98}{\%}. The discrepancy stems from many images containing background objects that are inconsistently reflected in their captions. We also find that the fraction generally decreases with the size of the pretraining set.

\paragraph{Overall Performance}
We find that CLIP's performance on the T2I task does not differ greatly between samples with known combinations (top row) and samples with novel combinations of known concepts (bottom row). The difference is slightly more pronounced on the I2T task, but performance is still high overall, indicating that CLIP generalizes well to novel object compositions.

\paragraph{Predicting Compositional Generalization}
We show a clear and consistent relationship between the average pretraining sample frequency $f_\text{avg}$ and CLIP's retrieval performance, even on samples requiring compositional generalization. The relation is approximately linear, except for the best-performing models, where it flattens as retrieval recall approaches 1.
Since the contribution of each object's pretraining frequency to the average pretraining sample frequency $f_\text{avg}$ is multiplicative, this consistent relationship implies that underrepresented objects are the bottleneck for compositional generalization.

\paragraph{Control on Synthetic Data} SynthCI-30M~\citep{hammoud2024synthclip} consists of synthetically generated images designed to cover a diverse combination of concepts.
Due to this process, we treat SynthCI-30 as a control to see if our results hold for a pretraining dataset sourced differently. We find that more test set combinations are unseen in SynthCI-30 than the pretraining sets derived from the real world, but the scaling of compositional generalization observed on real-world pretraining corpora also holds here.

Taken together, our results show that CLIP generalizes successfully to novel combinations of objects if it has observed the constituents sufficiently often during pretraining.
Note that objects in the pretraining data do not occur independently. In fact, many training samples contain multiple objects, and some object classes co-occur much more frequently than others.
The consistent scaling of CLIP's compositional generalization implies that the model can nonetheless disentangle objects and obtain an independent understanding of each object class.

For practitioners, our findings underline the importance of balancing object occurrences during data curation, as generalization is bottlenecked by the occurrence of each object.

\section{Next Steps}
\paragraph{Model Selection} While we control for architecture, parameter count, and pretraining scale, our experiments could be extended to other CLIP variants~\citep{yuksekgonul2023when}, diffusion models, or even LLMs.

\paragraph{Type of Compositionality} We only consider object compositions. We expect that our results may extend to attribute-object, foreground-background, texture-shape compositions in single-object scenes, since the independence assumption from Sec.~\ref{sec:f_sample} approximately holds. The bottleneck for these experiments is the concept-extraction pipeline. On the other hand, the scaling behavior of complex compositions, like attribute-binding with multiple objects, may not be as readily predictable.

\paragraph{Composition Granularity} Our definition of the co-occurence frequency in Eq.~\ref{eq:f_cap} only considers whether \emph{all} objects have jointly been observed during pretraining. For samples containing more than two objects, it might also be interesting to consider pair-wise object co-occurence and other partial co-occurences. How to integrate this information in the selection of test samples for compositional generalization remains is an open question.

\section{Conclusion}
Identifying conditions for successful real-world compositional generalization is a first step towards a future where models can be relied upon to generate new ideas, as \emph{``an idea is nothing more nor less than a new combination of old elements''}~\citep{young1975technique}. The ability to forecast when such capabilities will be unlocked is valuable not only to understand the compositional abilities of existing models but also to guide the development and scaling of future methods.
We take a first step in this direction by demonstrating how CLIP's ability to disentangle and recompose objects scales with the the frequency with which each object has been observed in the pretraining data.

\newpage

\subsection*{Acknowledgements}
We thank Vishaal Udandarao for helpful discussions and clarifying details regarding his work that we build our analysis on. We thank Rylan Schaeffer for helpful feedback on this manuscript. This work was supported by the German Federal Ministry of Education and Research (BMBF): Tübingen AI Center, FKZ: 01IS18039A. WB acknowledges financial support via an Emmy Noether Grant funded by the German Research Foundation (DFG) under grant no. BR 6382/1-1 and via the Open Philanthropy Foundation funded by the Good Ventures Foundation. WB is a Machine Learning Cluster of Excellence member, EXC number 2064/1 – Project number 390727645. This research utilized compute resources at the Tübingen Machine Learning Cloud, DFG FKZ INST 37/1057-1 FUGG.  MB and AP acknowledge financial support by the Federal Ministry of Education and Research (BMBF), FKZ: 011524085B and Open Philanthropy Foundation funded by the Good Ventures Foundation. We thank the International Max Planck Research School for Intelligent Systems (IMPRS-IS) for supporting TW.

\subsection*{Author Contributions}
The project was jointly led and coordinated by TW and YS. YS collected the data with input from TW and AP. TW and YS did the final analysis with input from AP. WB and MB participated in several helpful discussions during the project. TW, YS, and AP wrote the manuscript; TW made the figure with contributions from YS and AP.
{
    \small
    \bibliographystyle{plainnat}
    \bibliography{neurips_2024}

\begin{thebibliography}{43}
\providecommand{\natexlab}[1]{#1}
\providecommand{\url}[1]{\texttt{#1}}
\expandafter\ifx\csname urlstyle\endcsname\relax
  \providecommand{\doi}[1]{doi: #1}\else
  \providecommand{\doi}{doi: \begingroup \urlstyle{rm}\Url}\fi

\bibitem[Abbasi et~al.(2024)Abbasi, Rohban, and Baghshah]{abbasi2024deciphering}
Reza Abbasi, Mohammad~Hossein Rohban, and Mahdieh~Soleymani Baghshah.
\newblock Deciphering the role of representation disentanglement: Investigating compositional generalization in clip models.
\newblock In \emph{European Conference on Computer Vision (ECCV)}, 2024.

\bibitem[Antoniades et~al.(2024)Antoniades, Wang, Elazar, Amayuelas, Albalak, Zhang, and Wang]{antoniades2024generalization}
Antonis Antoniades, Xinyi Wang, Yanai Elazar, Alfonso Amayuelas, Alon Albalak, Kexun Zhang, and William~Yang Wang.
\newblock Generalization vs memorization: Tracing language models' capabilities back to pretraining data.
\newblock \emph{arXiv preprint arXiv:2407.14985}, 2024.

\bibitem[Berlot-Attwell et~al.(2023)Berlot-Attwell, Carrell, Agrawal, Sharma, and Saphra]{berlot2023attribute}
Ian Berlot-Attwell, A~Michael Carrell, Kumar~Krishna Agrawal, Yash Sharma, and Naomi Saphra.
\newblock Attribute diversity determines the systematicity gap in vqa.
\newblock \emph{arXiv preprint arXiv:2311.08695}, 2023.

\bibitem[Changpinyo et~al.(2021)Changpinyo, Sharma, Ding, and Soricut]{changpinyo2021conceptual}
Soravit Changpinyo, Piyush Sharma, Nan Ding, and Radu Soricut.
\newblock Conceptual 12m: Pushing web-scale image-text pre-training to recognize long-tail visual concepts.
\newblock In \emph{Conference on Computer Vision and Pattern Recognition (CVPR)}, 2021.

\bibitem[Dosovitskiy et~al.(2021)Dosovitskiy, Beyer, Kolesnikov, Weissenborn, Zhai, Unterthiner, Dehghani, Minderer, Heigold, Gelly, et~al.]{dosovitskiy2020image}
Alexey Dosovitskiy, Lucas Beyer, Alexander Kolesnikov, Dirk Weissenborn, Xiaohua Zhai, Thomas Unterthiner, Mostafa Dehghani, Matthias Minderer, Georg Heigold, Sylvain Gelly, et~al.
\newblock An image is worth 16x16 words: Transformers for image recognition at scale.
\newblock In \emph{International Conference on Learning Representations (ICLR)}, 2021.

\bibitem[Fodor and Pylyshyn(1988)]{fodor1988connectionism}
Jerry~A Fodor and Zenon~W Pylyshyn.
\newblock Connectionism and cognitive architecture: A critical analysis.
\newblock \emph{Cognition}, 28\penalty0 (1-2):\penalty0 3--71, 1988.

\bibitem[Goel et~al.(2022)Goel, Bansal, Bhatia, Rossi, Vinay, and Grover]{goel2022cyclip}
Shashank Goel, Hritik Bansal, Sumit Bhatia, Ryan Rossi, Vishwa Vinay, and Aditya Grover.
\newblock Cyclip: Cyclic contrastive language-image pretraining.
\newblock \emph{Advances in Neural Information Processing Systems}, 35:\penalty0 6704--6719, 2022.

\bibitem[Hammoud et~al.(2024)Hammoud, Itani, Pizzati, Torr, Bibi, and Ghanem]{hammoud2024synthclip}
Hasan Abed Al~Kader Hammoud, Hani Itani, Fabio Pizzati, Philip Torr, Adel Bibi, and Bernard Ghanem.
\newblock Synthclip: Are we ready for a fully synthetic clip training?
\newblock \emph{arXiv preprint arXiv:2402.01832}, 2024.

\bibitem[He et~al.(2016)He, Zhang, Ren, and Sun]{he2016deep}
Kaiming He, Xiangyu Zhang, Shaoqing Ren, and Jian Sun.
\newblock Deep residual learning for image recognition.
\newblock In \emph{Conference on Computer Vision and Pattern Recognition (CVPR)}, pages 770--778, 2016.

\bibitem[Hsieh et~al.(2023)Hsieh, Zhang, Ma, Kembhavi, and Krishna]{hsieh2023sugarcrepe}
Cheng-Yu Hsieh, Jieyu Zhang, Zixian Ma, Aniruddha Kembhavi, and Ranjay Krishna.
\newblock Sugarcrepe: Fixing hackable benchmarks for vision-language compositionality.
\newblock \emph{arXiv preprint arXiv:2306.14610}, 2023.

\bibitem[Huang et~al.(2023)Huang, Huang, Zhang, Tian, Feng, Zhang, Xie, Li, and Zhang]{huang2023open}
Xinyu Huang, Yi-Jie Huang, Youcai Zhang, Weiwei Tian, Rui Feng, Yuejie Zhang, Yanchun Xie, Yaqian Li, and Lei Zhang.
\newblock Open-set image tagging with multi-grained text supervision.
\newblock \emph{arXiv e-prints}, pages arXiv--2310, 2023.

\bibitem[Hupkes et~al.(2020)Hupkes, Dankers, Mul, and Bruni]{hupkes2020compositionality}
Dieuwke Hupkes, Verna Dankers, Mathijs Mul, and Elia Bruni.
\newblock Compositionality decomposed: How do neural networks generalise?
\newblock \emph{Journal of Artificial Intelligence Research}, 67:\penalty0 757--795, 2020.

\bibitem[Ilharco et~al.(2021)Ilharco, Wortsman, Wightman, Gordon, Carlini, Taori, Dave, Shankar, Namkoong, Miller, Hajishirzi, Farhadi, and Schmidt]{ilharco2021openclip}
Gabriel Ilharco, Mitchell Wortsman, Ross Wightman, Cade Gordon, Nicholas Carlini, Rohan Taori, Achal Dave, Vaishaal Shankar, Hongseok Namkoong, John Miller, Hannaneh Hajishirzi, Ali Farhadi, and Ludwig Schmidt.
\newblock Openclip, July 2021.
\newblock URL \url{https://doi.org/10.5281/zenodo.5143773}.

\bibitem[Jung et~al.(2024)Jung, Yoo, Ahn, and Hong]{jung2024learning}
Whie Jung, Jaehoon Yoo, Sungjin Ahn, and Seunghoon Hong.
\newblock Learning to compose: Improving object centric learning by injecting compositionality.
\newblock \emph{arXiv preprint arXiv:2405.00646}, 2024.

\bibitem[Kandpal et~al.(2023)Kandpal, Deng, Roberts, Wallace, and Raffel]{kandpal2023large}
Nikhil Kandpal, Haikang Deng, Adam Roberts, Eric Wallace, and Colin Raffel.
\newblock Large language models struggle to learn long-tail knowledge.
\newblock In \emph{International Conference on Machine Learning (ICML)}, pages 15696--15707. PMLR, 2023.

\bibitem[Kim and Linzen(2020)]{kim2020cogs}
Najoung Kim and Tal Linzen.
\newblock Cogs: A compositional generalization challenge based on semantic interpretation.
\newblock In \emph{Conference on Empirical Methods in Natural Language Processing (EMNLP)}, pages 9087--9105, 2020.

\bibitem[Lake and Baroni(2018)]{lake2018generalization}
Brenden Lake and Marco Baroni.
\newblock Generalization without systematicity: On the compositional skills of sequence-to-sequence recurrent networks.
\newblock In \emph{International Conference on Machine Learning (ICML)}, 2018.

\bibitem[Lake and Baroni(2023)]{lake2023human}
Brenden~M Lake and Marco Baroni.
\newblock Human-like systematic generalization through a meta-learning neural network.
\newblock \emph{Nature}, 623\penalty0 (7985):\penalty0 115--121, 2023.

\bibitem[Lewis et~al.(2024)Lewis, Nayak, Yu, Yu, Merullo, Bach, and Pavlick]{lewis2022does}
Martha Lewis, Nihal~V Nayak, Peilin Yu, Qinan Yu, Jack Merullo, Stephen~H Bach, and Ellie Pavlick.
\newblock Does clip bind concepts? probing compositionality in large image models.
\newblock \emph{European Chapter of the Association for Computational Linguistics (EACL)}, 2024.

\bibitem[Lin et~al.(2014)Lin, Maire, Belongie, Hays, Perona, Ramanan, Doll{\'a}r, and Zitnick]{lin2014microsoft}
Tsung-Yi Lin, Michael Maire, Serge Belongie, James Hays, Pietro Perona, Deva Ramanan, Piotr Doll{\'a}r, and C~Lawrence Zitnick.
\newblock Microsoft coco: Common objects in context.
\newblock In \emph{European Conference on Computer Vision (ECCV)}, 2014.

\bibitem[Ma et~al.(2023)Ma, Hong, Gul, Gandhi, Gao, and Krishna]{ma2023crepe}
Zixian Ma, Jerry Hong, Mustafa~Omer Gul, Mona Gandhi, Irena Gao, and Ranjay Krishna.
\newblock Crepe: Can vision-language foundation models reason compositionally?
\newblock In \emph{Proceedings of the IEEE/CVF Conference on Computer Vision and Pattern Recognition}, pages 10910--10921, 2023.

\bibitem[Mayilvahanan et~al.(2024)Mayilvahanan, Zimmermann, Wiedemer, Rusak, Juhos, Bethge, and Brendel]{mayilvahanan2024search}
Prasanna Mayilvahanan, Roland~S Zimmermann, Thadd{\"a}us Wiedemer, Evgenia Rusak, Attila Juhos, Matthias Bethge, and Wieland Brendel.
\newblock In search of forgotten domain generalization.
\newblock In \emph{ICML 2024 Workshop on Foundation Models in the Wild}, 2024.

\bibitem[Montero et~al.(2022)Montero, Bowers, Costa, Ludwig, and Malhotra]{monteroLostLatentSpace2022}
Milton~L. Montero, Jeffrey Bowers, Rui~Ponte Costa, Casimir~JH Ludwig, and Gaurav Malhotra.
\newblock Lost in {{Latent Space}}: {{Examining}} failures of disentangled models at combinatorial generalisation.
\newblock In \emph{Advances in {{Neural Information Processing Systems}}}, October 2022.

\bibitem[Montero et~al.(2021)Montero, Ludwig, Costa, Malhotra, and Bowers]{montero2021role}
Milton~Llera Montero, Casimir~JH Ludwig, Rui~Ponte Costa, Gaurav Malhotra, and Jeffrey Bowers.
\newblock The role of disentanglement in generalisation.
\newblock In \emph{International Conference on Learning Representations (ICLR)}, 2021.

\bibitem[Mu et~al.(2022)Mu, Kirillov, Wagner, and Xie]{mu2022slip}
Norman Mu, Alexander Kirillov, David Wagner, and Saining Xie.
\newblock Slip: Self-supervision meets language-image pre-training.
\newblock In \emph{European Conference on Computer Vision (ECCV)}, 2022.

\bibitem[Nguyen et~al.(2022)Nguyen, Ilharco, Wortsman, Oh, and Schmidt]{nguyen2022quality}
Thao Nguyen, Gabriel Ilharco, Mitchell Wortsman, Sewoong Oh, and Ludwig Schmidt.
\newblock Quality not quantity: On the interaction between dataset design and robustness of clip.
\newblock \emph{Advances in Neural Information Processing Systems}, 35:\penalty0 21455--21469, 2022.

\bibitem[Okawa et~al.(2023)Okawa, Lubana, Dick, and Tanaka]{okawaCompositionalAbilitiesEmerge2023}
Maya Okawa, Ekdeep~Singh Lubana, Robert~P. Dick, and Hidenori Tanaka.
\newblock Compositional {{Abilities Emerge Multiplicatively}}: {{Exploring Diffusion Models}} on a {{Synthetic Task}}.
\newblock June 2023.

\bibitem[Radford et~al.(2021)Radford, Kim, Hallacy, Ramesh, Goh, Agarwal, Sastry, Askell, Mishkin, Clark, et~al.]{radford2021learning}
Alec Radford, Jong~Wook Kim, Chris Hallacy, Aditya Ramesh, Gabriel Goh, Sandhini Agarwal, Girish Sastry, Amanda Askell, Pamela Mishkin, Jack Clark, et~al.
\newblock Learning transferable visual models from natural language supervision.
\newblock In \emph{International Conference on Machine Learning (ICML)}, 2021.

\bibitem[Ray et~al.(2024)Ray, Radenovic, Dubey, Plummer, Krishna, and Saenko]{ray2024cola}
Arijit Ray, Filip Radenovic, Abhimanyu Dubey, Bryan Plummer, Ranjay Krishna, and Kate Saenko.
\newblock Cola: A benchmark for compositional text-to-image retrieval.
\newblock \emph{Advances in Neural Information Processing Systems}, 36, 2024.

\bibitem[Schott et~al.(2022)Schott, Von~K{\"u}gelgen, Tr{\"a}uble, Gehler, Russell, Bethge, Sch{\"o}lkopf, Locatello, and Brendel]{schott2022visual}
Lukas Schott, Julius Von~K{\"u}gelgen, Frederik Tr{\"a}uble, Peter~Vincent Gehler, Chris Russell, Matthias Bethge, Bernhard Sch{\"o}lkopf, Francesco Locatello, and Wieland Brendel.
\newblock Visual representation learning does not generalize strongly within the same domain.
\newblock In \emph{International Conference on Learning Representations}, 2022.

\bibitem[Schuhmann et~al.(2021)Schuhmann, Vencu, Beaumont, Kaczmarczyk, Mullis, Katta, Coombes, Jitsev, and Komatsuzaki]{schuhmann2021laion}
Christoph Schuhmann, Richard Vencu, Romain Beaumont, Robert Kaczmarczyk, Clayton Mullis, Aarush Katta, Theo Coombes, Jenia Jitsev, and Aran Komatsuzaki.
\newblock Laion-400m: Open dataset of clip-filtered 400 million image-text pairs.
\newblock \emph{arXiv preprint arXiv:2111.02114}, 2021.

\bibitem[Sharma et~al.(2018)Sharma, Ding, Goodman, and Soricut]{sharma2018conceptual}
Piyush Sharma, Nan Ding, Sebastian Goodman, and Radu Soricut.
\newblock Conceptual captions: A cleaned, hypernymed, image alt-text dataset for automatic image captioning.
\newblock In \emph{Proceedings of the 56th Annual Meeting of the Association for Computational Linguistics (Volume 1: Long Papers)}, pages 2556--2565, 2018.

\bibitem[Sun et~al.(2023)Sun, Williams, and Hupkes]{sun-etal-2023-validity}
Kaiser Sun, Adina Williams, and Dieuwke Hupkes.
\newblock The validity of evaluation results: Assessing concurrence across compositionality benchmarks.
\newblock In Jing Jiang, David Reitter, and Shumin Deng, editors, \emph{Proceedings of the 27th Conference on Computational Natural Language Learning (CoNLL)}, pages 274--293, Singapore, December 2023. Association for Computational Linguistics.
\newblock \doi{10.18653/v1/2023.conll-1.19}.
\newblock URL \url{https://aclanthology.org/2023.conll-1.19}.

\bibitem[Thomee et~al.(2016)Thomee, Shamma, Friedland, Elizalde, Ni, Poland, Borth, and Li]{thomee2016yfcc100m}
Bart Thomee, David~A Shamma, Gerald Friedland, Benjamin Elizalde, Karl Ni, Douglas Poland, Damian Borth, and Li-Jia Li.
\newblock Yfcc100m: The new data in multimedia research.
\newblock \emph{Communications of the ACM}, 59\penalty0 (2):\penalty0 64--73, 2016.

\bibitem[Thrush et~al.(2022)Thrush, Jiang, Bartolo, Singh, Williams, Kiela, and Ross]{thrush2022winoground}
Tristan Thrush, Ryan Jiang, Max Bartolo, Amanpreet Singh, Adina Williams, Douwe Kiela, and Candace Ross.
\newblock Winoground: Probing vision and language models for visio-linguistic compositionality.
\newblock In \emph{Conference on Computer Vision and Pattern Recognition (CVPR)}, 2022.

\bibitem[Udandarao et~al.(2024)Udandarao, Prabhu, Ghosh, Sharma, Torr, Bibi, Albanie, and Bethge]{udandarao2024zeroshotexponentialdatapretraining}
Vishaal Udandarao, Ameya Prabhu, Adhiraj Ghosh, Yash Sharma, Philip H.~S. Torr, Adel Bibi, Samuel Albanie, and Matthias Bethge.
\newblock No "zero-shot" without exponential data: Pretraining concept frequency determines multimodal model performance, 2024.
\newblock URL \url{https://arxiv.org/abs/2404.04125}.

\bibitem[Wang et~al.(2024)Wang, Si, Shao, and Zhao]{wangEnhancingCompositionalGeneralization2024}
Haoxiang Wang, Haozhe Si, Huajie Shao, and Han Zhao.
\newblock Enhancing {{Compositional Generalization}} via {{Compositional Feature Alignment}}, February 2024.

\bibitem[Wiedemer et~al.(2023)Wiedemer, Mayilvahanan, Bethge, and Brendel]{wiedemer2023compositional}
Thadd\"{a}us Wiedemer, Prasanna Mayilvahanan, Matthias Bethge, and Wieland Brendel.
\newblock Compositional generalization from first principles.
\newblock In A.~Oh, T.~Naumann, A.~Globerson, K.~Saenko, M.~Hardt, and S.~Levine, editors, \emph{Advances in Neural Information Processing Systems}, volume~36, pages 6941--6960. Curran Associates, Inc., 2023.
\newblock URL \url{https://proceedings.neurips.cc/paper_files/paper/2023/file/15f6a10899f557ce53fe39939af6f930-Paper-Conference.pdf}.

\bibitem[Wiedemer et~al.(2024)Wiedemer, Brady, Panfilov, Juhos, Bethge, and Brendel]{wiedemer2024provable}
Thadd{\"a}us Wiedemer, Jack Brady, Alexander Panfilov, Attila Juhos, Matthias Bethge, and Wieland Brendel.
\newblock Provable compositional generalization for object-centric learning.
\newblock In \emph{The Twelfth International Conference on Learning Representations}, 2024.
\newblock URL \url{https://openreview.net/forum?id=7VPTUWkiDQ}.

\bibitem[Young and Reinhard(1975)]{young1975technique}
James~Webb Young and Keith Reinhard.
\newblock \emph{A technique for producing ideas}.
\newblock NTC Business Books, 1975.

\bibitem[Young et~al.(2014)Young, Lai, Hodosh, and Hockenmaier]{young2014image}
Peter Young, Alice Lai, Micah Hodosh, and Julia Hockenmaier.
\newblock From image descriptions to visual denotations: New similarity metrics for semantic inference over event descriptions.
\newblock \emph{Transactions of the Association for Computational Linguistics}, 2:\penalty0 67--78, 2014.

\bibitem[Yuksekgonul et~al.(2023)Yuksekgonul, Bianchi, Kalluri, Jurafsky, and Zou]{yuksekgonul2023when}
Mert Yuksekgonul, Federico Bianchi, Pratyusha Kalluri, Dan Jurafsky, and James Zou.
\newblock When and why vision-language models behave like bags-of-words, and what to do about it?
\newblock In \emph{The Eleventh International Conference on Learning Representations}, 2023.
\newblock URL \url{https://openreview.net/forum?id=KRLUvxh8uaX}.

\bibitem[Zhao et~al.(2022)Zhao, Zhang, Zhu, Shen, Lee, Lu, and Yin]{zhao2022vl}
Tiancheng Zhao, Tianqi Zhang, Mingwei Zhu, Haozhan Shen, Kyusong Lee, Xiaopeng Lu, and Jianwei Yin.
\newblock Vl-checklist: Evaluating pre-trained vision-language models with objects, attributes and relations.
\newblock \emph{arXiv preprint arXiv:2207.00221}, 2022.

\end{thebibliography}
}


\newpage
\appendix
\section{Additional Evaluations}\label{app:extra_plots}
\begin{figure}[h!]
    \centering
    \includegraphics[width=\linewidth]{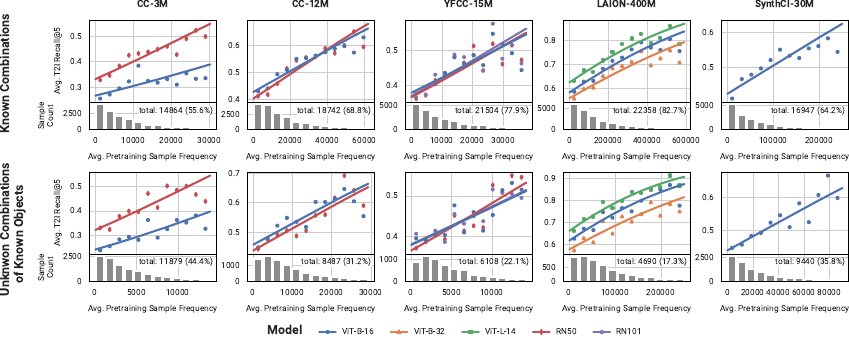}
    \caption{\textbf{T2I Recall@5} We see that on combinations that are both known and unknown to the model, across architectures and pretraining sets, there exists a predictive relationship between the sample frequency, i.e. the aggregated frequencies of objects in the combination, and the performance.}
    \label{fig:lines_t2i_k5}
\end{figure}

\begin{figure}[h!]
    \centering
    \includegraphics[width=\linewidth]{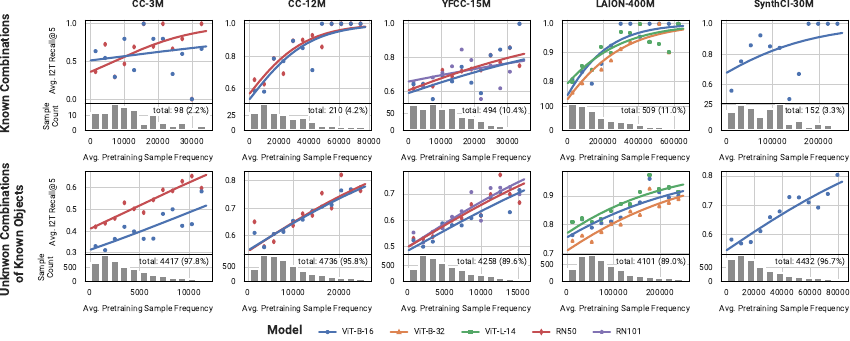}
    \caption{\textbf{I2T Recall@5} We see that on combinations that are both known and unknown to the model, across architectures and pretraining sets, there exists a predictive relationship between the sample frequency, i.e. the aggregated frequencies of objects in the combination, and the performance.}
    \label{fig:lines_i2t_k5}
\end{figure}

\begin{figure}[h!]
    \centering
    \includegraphics[width=\linewidth]{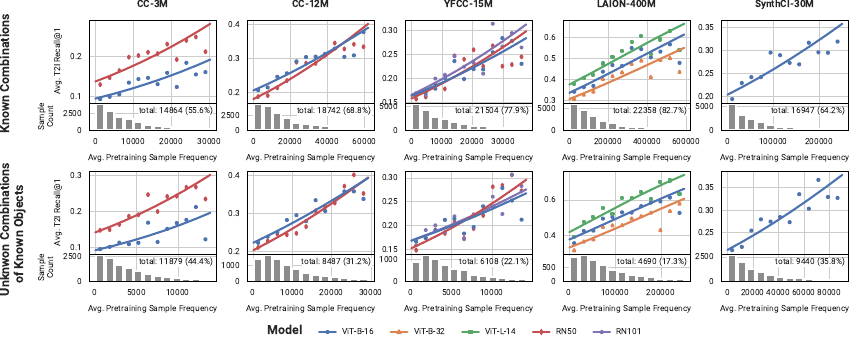}
    \caption{\textbf{T2I Recall@1} We see that on combinations that are both known and unknown to the model, across architectures and pretraining sets, there exists a predictive relationship between the sample frequency, i.e. the aggregated frequencies of objects in the combination, and the performance.}
    \label{fig:lines_t2i_k1}
\end{figure}

\begin{figure}[h!]
    \centering
    \includegraphics[width=\linewidth]{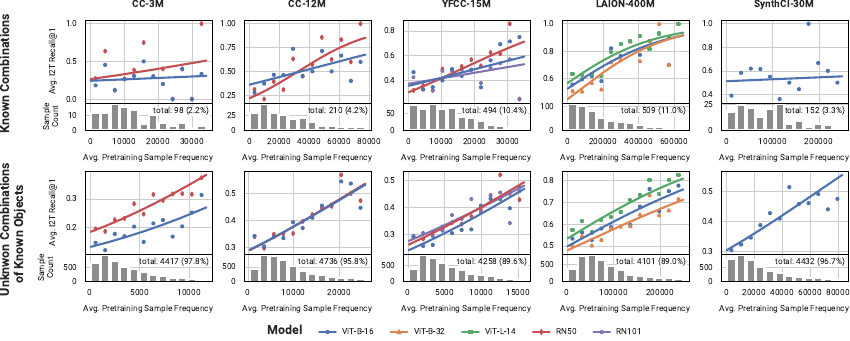}
    \caption{\textbf{I2T Recall@1} We see that on combinations that are both known and unknown to the model, across architectures and pretraining sets, there exists a predictive relationship between the sample frequency, i.e. the aggregated frequencies of objects in the combination, and the performance.}
    \label{fig:lines_i2t_k1}
\end{figure}

~\Cref{fig:lines_t2i_k5,fig:lines_i2t_k5,fig:lines_t2i_k1,fig:lines_i2t_k1} plot trends seen in~\Cref{fig:lines_t2i_k10,fig:lines_i2t_k10} for Recall@5 and Recall@1. We observe similar trends.

\end{document}